# Recovering Articulated Object Models from 3D Range Data

**Dragomir Anguelov**     **Daphne Koller**
**Hoi-Cheung Pang**     **Praveen Srinivasan**     **Sebastian Thrun**
Computer Science Dept.
Stanford University
Stanford, CA 94305
{*drago,koller,hcpang,praveens,thrun*}@cs.stanford.edu

## Abstract

We address the problem of unsupervised learning of complex articulated object models from 3D range data. We describe an algorithm whose input is a set of meshes corresponding to different configurations of an articulated object. The algorithm automatically recovers a decomposition of the object into approximately rigid parts, the location of the parts in the different object instances, and the articulated object skeleton linking the parts. Our algorithm first registers all the meshes using an unsupervised non-rigid technique described in a companion paper. It then segments the meshes using a graphical model that captures the spatial contiguity of parts. The segmentation is done using the EM algorithm, iterating between finding a decomposition of the object into rigid parts, and finding the location of the parts in the object instances. Although the graphical model is densely connected, the object decomposition step can be performed optimally and efficiently, allowing us to identify a large number of object parts while avoiding local maxima. We demonstrate the algorithm on real world datasets, recovering a 15-part articulated model of a human puppet from just 7 different puppet configurations, as well as a 4 part model of a flexing arm where significant non-rigid deformation was present.

## 1 Introduction

Articulated objects consist of approximately rigid parts, which are linked by joints to form an object skeleton; examples include the human body, most animals, office chairs, cars and many others. Models of such articulated bodies have been used extensively in computer vision and in graphics, for applications such as object pose detection and tracking in video [16, 7, 28, 23] and in 3D data [20], and for 3D motion estimation and rendering [1]. In the vast majority of applications, the articulated skeleton structure and parameters are specified by hand, although some applications optimize the parameters of the individual joints [1, 21, 14]. An algorithm that recovers articulate models from 3D data in a completely unsupervised way can greatly decrease the dependence in many of these applications on human-engineered articulated models.

In this article, we propose an algorithm that takes a set of meshes corresponding to different configurations of an articulated object as input and recovers the articulated skeleton of the object. The algorithm first registers the input meshes using an unsupervised non-rigid registration algorithm — the Correlated Correspondences algorithm [2]. Taking the resulting mesh registrations as given, we define a graphical model capturing the structure of the part decomposition problem. For each mesh point, the model defines a hidden variable determining which object part the point belongs to. Reflecting our premise that the object is largely rigid with articulate parts, out model assumes that all of the points in each part move in the same way, that is, they all undergo a single rigid transformation in each of the registered meshes. Finally, in order to ensure that our part decomposition is reasonable, our model also includes soft spatial contiguity constraints, encoding a preference for part decompositions where nearby mesh points are assigned to the same part; these constraints are encoded as undirected edges in the graphical model. Our algorithm then performs *Expectation-Maximization* on the resulting model, iterating between finding a decomposition of the object into rigid parts, and finding the location of the parts in the object instances. Despite the fact that the underlying graphical model is densely connected, we show how the object decomposition into parts can be performed as a global optimization step, which allows us to identify a large number of object parts while avoiding local maxima. Given a decomposition of the object into rigid parts, we show how to estimate the articulated skeleton linking the parts.

We tested our algorithm on two real-world datasets, where it achieved very good results. We demonstrate automatic recovery of a 15-part articulated model of a human puppet from just 7 different configurations of the puppet. To our knowledge, this is the first implementation that estimates such a complex skeleton from real world data from very few poses, in a completely unsupervised way. We also demonstrate that the algorithm performs well in the presence of significant non-rigid motion, by demonstrating automatic recovery of a 4-part model of a human arm.

## 2  Related Work

Our algorithm is related to a number of clustering approaches, which attempt to partition the input from a scene into several coherent regions.

Early work in vision directly clusters motion estimates obtained by locally tracking 2D image patches in video data and representing the scene as a set of image layers [26, 4]. These approaches assume small local motion and do not readily generalize to 3D data. They do not address articulation because the layers are allowed to have arbitrary shape and connectivity.

Recently, some approaches on finding articulate models in 2D data were proposed as well. The work of Song *et al.* [24] demonstrates recovery of articulated human models represented as decomposable triangulated graphs from tracked 2D features in video. Decomposable triangulated graphs are a limited class of graphs, unsuitable for representing 3D shapes and articulation in 3D. Additionally, recovering articulation in 2D is considerably more challenging, due to the information loss arising from the projection of a 3D scene to 2D. As a consequence, the models recovered using such procedures tend to be very sparse (containing about a dozen points and triangles), and are fairly far from being realistic human models [24].

Another class of related methods use non-negative matrix factorization to decompose a database of images into a set of parts [18, 12]. These methods treat each image as an additive collection of a set of basis images, which turn out to be rather sparse due to the non-negativity constraints. However, such an additive model is not suitable for modeling articulation, because it does not attempt to associate parts in one image with parts in another, and in particular treats the same arm in raised and lowered position as two different "parts". This method was also applied only to 2D data, and the extension to 3D does not appear obvious.

The work of Taycher *et al.* [25], which applies in both 2D and 3D, shows how to recover tree-shaped articulation models, but it makes the restrictive assumption that the rigid parts and their transformations are known.

Our approach is most directly related to the work of Cheung *et al.* [8], which shows how to estimate articulated object models from 3D Shape-From-Silhouette (SFS) data, augmented with information about object color. They report recovering an articulated human model with 9 parts. However, their algorithm was applied only to sequences where a single body part is moving at a time. Each sequence contains two articulated parts, and allows the estimation of a single human joint. The final articulated model is generated by combining the joints estimated in all the two-part sequences. Despite the important cue of color information, they did not demonstrate simultaneous recovery of multiple parts.

We believe that the reason for this limitation is the presence of local maxima in their approach, arising for two reasons. First, they solve for the point correspondences between the input meshes while solving for the articulated model. The approach is a generalization of the *Iterative Closest Point (ICP)* algorithm [6] to multiple rigid parts. However, ICP is known to be prone to local maxima (see a discussion of the problems with ICP in [2]). The additional degrees of freedom provided by the possible part decompositions make the problem more severe. By contrast, we take a two-phase approach, first solving the correspondence problem using a non-rigid registration technique that allows large deformations, and then learning the articulated object model. Our approach has the potential limitation of ignoring information about coherent part motion in solving the registration problem. Nevertheless, its ability to circumvent many local maxima appears to significantly offset that potential disadvantage.

A second source for local maxima arises from the choice of constraints enforcing that parts are contiguous regions of the object surface. The approach of Cheung *et al.* enforces part contiguity with discrete constraints between assignments to mesh points and their neighbors. This type of model does not allow the application of efficient global optimization steps. By contrast, our algorithm enforces part contiguity using soft probabilistic constraints, which allow us to violate these constraints locally as long as it is maximizing the log-likelihood of the model as a whole. Moreover, we can apply efficient global optimization methods to determine the optimal part decomposition. These two properties allow us to be less sensitive to initialization, and to avoid local maxima even if a large number of parts is present.

## 3  Mesh Registration

We start with a set of meshes $D_0 \ldots D_N$ corresponding to different configurations of the same object. We assume that the object is composed of a number of rigid parts, whose orientations may vary in each configuration. Each mesh $D_i$ is a tesselation of the respective surface into polygons (usually triangles), and contains the polygon vertices and the links between them.

We pick one of the meshes to be a *template mesh* $X \equiv D_0$ and denote its set of polygon edges as $E(X)$. We automatically register this template $X$ with the remaining meshes $D_1, \ldots, D_n$. The registration process aligns the template $X$ with each mesh $D_i$ producing a transformed *instance mesh* $Z_i$ in each case. Each instance mesh $Z_i$ is produced by applying a non-rigid transformation to the points of $X$, so it has the same set of points and edges as $X$. Moreover, the one-to-one correspondences between the instance mesh and the template mesh points and edges are known. Throughout our discussion, we will use the fact that a particular point $x_j$ of mesh $X$ corresponds to point $z_{i,j}$ in mesh $Z_i$.

We use a non-rigid registration algorithm called Correlated Correspondences (an independent contribution described in a companion tech report [2]), which can handle large deformations of the mesh surface. The Correlated Correspondence algorithm formulates the registration problem as one of finding a deformable embedding of one mesh into another. It is related to work in the vision community on deformable template matching [9, 13]. The algorithm finds the registration between points in the two meshes by using a joint probabilistic model over all point-to-point correspondences. The model encodes the correlations between the correspondence variable assignments, which enforce the preservation of local geometry, and the preservation of geodesic distance between corresponding pairs of points in the two meshes. *Loopy belief propagation (LBP)* [27] is used to find a good joint assignment to all correspondence variables, which defines the deformable embedding. The registration can be additionally refined by applying EM-style iterative techniques [15, 22].

The Correlated Correspondence algorithm does not use markers, nor does it assume prior knowledge about object shape, the dynamics of its deformation, or the initial alignment of meshes. It successfully registers scans that exhibit large transformations, including both movement of articulate parts and non-rigid surface deformations. The Correlated Correspondence algorithm is not universally applicable to any two scans, because of its assumption that local geodesic distances are approximately preserved. In cases when mesh topology changes significantly, for example when an arm touches the head, it may fail.

The algorithm for recovering articulated models which we present in this paper is independent of the Correlated Correspondence algorithm and is appropriate whenever a reasonable registration of several scans is available.

## 4 Probabilistic Framework

### 4.1 Generating the Instance Meshes

In this section we describe our beliefs about the process which transforms the template mesh $X$ into the instance meshes $Z_1, \ldots, Z_N$.

We assume that the surface of $X$ is made up of the set of $\mathcal{P} = \{1, \ldots, P\}$ rigid parts. We associate each template mesh point $x_j$ with a part label $\alpha_j$, denoting the rigid part to which the point belongs. Each label $\alpha_j$ can take one of $P$ possible values.

Every rigid part $p$ is associated with a set of transformations $T_{1,p}, \ldots, T_{N,p}$, one for each instance mesh. All points assigned to part $p$ share this set of transformations. More precisely, $\tilde{x}_{i,j} = T_{i,\alpha_j}(x_j)$, where $\tilde{x}_{i,j}$ denotes the transformed location of $x_j$ in instance $i$. We want to model objects which are not perfectly rigid, so we allow the point locations $z_{i,j}$ in the meshes $Z_i$ to deviate from these predicted locations. We assume that each point location $z_{i,j}$ is generated from $\tilde{x}_{i,j}$ by a Gaussian process:

$$P(z_{i,j} \mid \alpha_j = p, T_{i,p}) = \mathcal{N}(z_{i,j}; \tilde{x}_{i,j}, \mathbf{diag}(\sigma^2)) \quad (1)$$

where $\sigma^2$ is the variance, chosen to be a multiple of the resolution of mesh $X$.

### 4.2 Introducing Contiguity Constraints

So far, our model allows a part to be composed of an arbitrary set of points interspersed throughout the mesh. What we actually want is that each part is comprised of a set of points in a connected region.

We choose to enforce this preference by two kinds of constraints. *Soft contiguity constraints* enforce the preference that neighboring points in the template mesh have similar part labels. More formally, we define two labels $\alpha_j$ and $\alpha_k$ to be neighboring if their corresponding points $x_j$ and $x_k$ are connected by an edge in $X$. We model soft contiguity constraints as probabilistic potentials between all neighboring pairs of labels $\alpha_j$ and $\alpha_k$:

$$\phi(\alpha_j, \alpha_k) = \left\{ \begin{array}{rcl} \tau & : & \alpha_j = \alpha_k \\ 1 - \tau & : & \alpha_j \neq \alpha_k \end{array} \right. \quad (2)$$

where $\tau > 0.5$.

These soft constraints bias us toward a partitioning of the template mesh into contiguous regions. They induce a probabilistic model which can be optimized efficiently (as we will shortly discuss). However, soft contiguity constraints still allow each part to be comprised of several disjoint components. For example, if the arms of an office chair always get raised and lowered together, the model as stated above will allow both arms to belong to the same object part. Such models can be preferable in some situations but they are not appropriate for recovering an articulated object skeleton: the notion of a joint between parts is not well-defined when each part consists of several disconnected regions. In order to model the object articulation correctly, we impose another kind of constraint, which we call *hard contiguity constraint*. The constraint specifies that a part can consist of no more than one connected component in the template mesh.

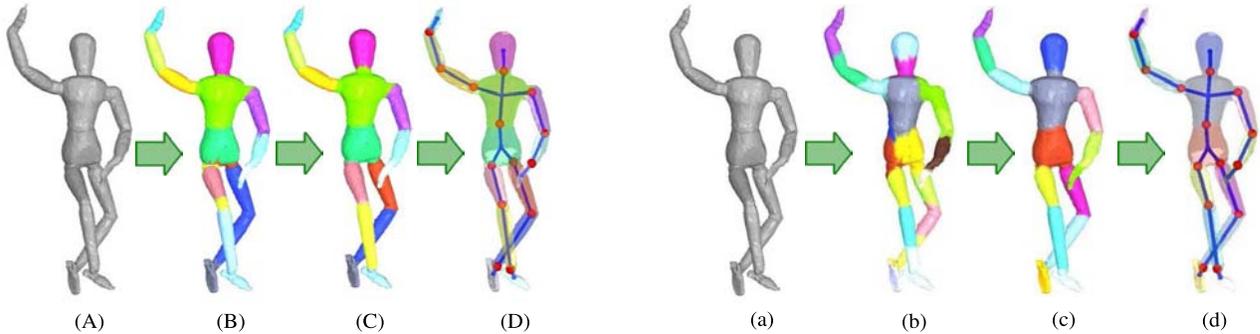

Figure 1: Illustration of the part-finding process: (A),(a) a template mesh is registered to all other meshes by CC algorithm. (B) the mesh is divided into parts by clustering the estimated local transformations for each template point, different parts are color-coded. (b) the mesh is randomly divided into small patches of approximately equal areas, different parts are color-coded. (C),(c) results in (B),(b) are used to initialize the EM algorithm which solves for the part assignments and the transformation for each part. (D),(d) the joints linking the rigid parts are estimated.

### 4.3 Model Summary

Ignoring the hard contiguity constraints, the framework described in Sec. 4 defines a *Markov network* over the part labels $\alpha$. A Markov network encodes the joint distribution over a set of variables as a product of potentials:

$$P(\alpha) = \frac{1}{Z} \prod_j \phi(\alpha_j) \prod_{j,k} \phi(\alpha_j, \alpha_k) \qquad (3)$$

where $Z$ is a normalization constant.

The singleton potentials $\phi(\alpha_j)$ correspond to the probabilities that a template point $x_j$ generates its corresponding points $z_{1,j}, \ldots, z_{N,j}$, as follows:

$$\phi(\alpha_j = p) = \prod_{i=1}^N P(z_{i,j} \mid \alpha_j = p, T_{i,p}) \qquad (4)$$

The potential values depend on the transformations $T_{i,p}$. Thus, the joint distribution depends on $T$, the set of rigid part transformations. The pairwise potentials in the Markov network correspond to the soft contiguity constraints, and are defined in Eq. (2).

## 5 Optimization

We start with a template mesh $X$ and instance meshes $Z_1, \ldots, Z_N$, and we need to solve for the set of part transformations $T$, as well as for the part labels $\alpha$.

We want to find a joint assignment to the part labels and the transformations which maximizes the log-likelihood of the model:

$$\operatorname*{argmax}_{\alpha,T} \log P(\alpha, T) = \operatorname*{argmax}_{\alpha,T} \{ \sum_{(j,k) \in E(X)} \log \phi(\alpha_j, \alpha_k) - \frac{1}{2\sigma^2} \sum_{i=1}^n \sum_{j=1}^J \|z_{i,j} - T_{i,\alpha_j}(x_j)\|^2 \} \quad (5)$$

where $J$ is the number of points in meshes $X, Z_1, \ldots, Z_N$. Note that our objective is defined as optimizing both the part assignment and transformations simultaneously, rather than marginalizing over the (hidden) part assignment variables. A hard assignment of points into parts is very appropriate for our application, and it also allows the use of efficient global optimization steps, as we discuss below. Note that the hard contiguity constraints are not accounted for in the above equation, and have to be enforced separately.

The objective in Eq. (5) is non-convex in the set of variables $\alpha, T$. We optimize it using hard *Expectation-Minimization (EM)* to find a good assignment for $\alpha, T$ in an iterative fashion. EM iterates between two steps: the *E-step* calculates a hard assignment for all part labels $\alpha$ given an estimate of the transformations $T$. The *M-step* improves the estimate for the parameters $T$ using the labels $\alpha$ obtained in the E-step.

### 5.1 E-Step

Our goal in the E-step is to find the MAP assignment to the part labels maximizing Eq. (5) for a given set of transformations $T$. It turns out that this is an instance of the Uniform Labeling problem [17], which can be expressed as an integer program. Following Kleinberg and Tardos [17], we introduce indicator variables $\alpha_{jp}$ for the event $\alpha_j = p$, where we require that for all $p$ $\alpha_{jp} \in \{0,1\}$ and $\sum_{p=1}^{P} \alpha_{jp} = 1$. These integer constraints imply that we have only a single $p$ for which $\alpha_{jp} = 1$, and the others are all 0. The log-cost associated with a particular single potential can then be expressed as $\sum_{p=1}^{P} c(j,p)\alpha_{jp}$ where

$$c(j,p) = -\frac{1}{2\sigma^2} \sum_{i=1}^{N} \|z_{i,j} - T_{i,p}(x_j)\|^2 \qquad (6)$$

The *separation cost* of an edge in mesh $X$ can also be defined in terms of the variables $\alpha_{jp}$. The difference between the labels of the edge endpoints can be expressed as

$$\beta_{jk} = \frac{1}{2} \sum_{p=1}^{P} |\alpha_{jp} - \alpha_{kp}| = \frac{1}{2} \sum_{p=1}^{P} \beta_{jkp} \qquad (7)$$

where $\beta_{jkp} = |\alpha_{jp} - \alpha_{kp}|$. The cost associated with an edge is therefore $s \cdot \beta_{jk}$, where $s = \log(\tau) - \log(1-\tau)$.

We can now rewrite our optimization problem as an integer program:

$$\max \sum_{j=1}^{J} \sum_{p=1}^{P} c(j,p) \cdot \alpha_{jp} + \sum_{(j,k) \in E(X)} s \cdot \beta_{j,k}$$

$$\text{s.t.} \sum_{p=1}^{P} \alpha_{jp} = 1, \qquad j = \{1,\ldots,J\}$$

$$\beta_{jk} = \frac{1}{2} \sum_{p=1}^{P} \beta_{jkp}, \ (j,k) \in E(X),$$

$$\beta_{jkp} \geq \alpha_{jp} - \alpha_{kp}, \ (j,k) \in E(X), p \in \mathcal{P}$$

$$\beta_{jkp} \geq \alpha_{kp} - \alpha_{jp}, \ (j,k) \in E(X), p \in \mathcal{P}$$

$$\alpha_{jp} \in \{0,1\}, \qquad j = \{1,\ldots,J\}, p \in \mathcal{P}$$

In general, solving an integer program optimally is NP-hard. However, we can define a linear programming relaxation of the above problem by replacing the integrality constraints $\alpha_{jp} \in \{0,1\}$ with $\alpha_{jp} \geq 0$. This relaxation allows fractional solutions for the labels $\alpha_j$. The linear program can be solved very efficiently by a solver such as CPLEX.

For problems of this type, Kleinberg and Tardos [17] describe a method for rounding the fractional solution, losing at most a factor of 2 in the objective function. In our experiments we did not need to perform this rounding because the relaxed linear formulation always returned integer solutions. In this case, we are guaranteed that our solution is the optimal assignment of template mesh points to parts, which maximizes Eq. (5) given a set of rigid transformations $T$.

The assignment described so far does not enforce the hard contiguity constraints. However, we can easily test if one of the parts returned by this solution is split into several disconnected components in the mesh. In this case, we simply split the part into its connected components, introducing a separate part for each one. This step satisfies the hard contiguity constraints, while preserving the value of the objective function in Eq. (5).

### 5.2 M-Step

The goal of the M-step is to find the set of rigid part transformations $T$ which maximize the log-likelihood in Eq. (5), given the part label assignments $\alpha$ supplied by the E-step. The objective function decomposes into a separate equation for each $T_{i,p}$:

$$\underset{T_{i,p}}{\operatorname{argmin}} \sum_{j=1}^{J} I(\alpha_j = p) \cdot \|z_{ij} - T_{i,p}(x_j)\|^2 \qquad (8)$$

where $I(\cdot)$ is the indicator function. This problem is isomorphic to the registration problem studied extensively in the ICP literature. We adapt the canonical solution to this problem, proposed by Besl and Mckay [6], where 3D rotations are represented as quaternions, and a closed form estimate of $T_{i,p}$ is obtained by solving a small system of linear equations.

## 6 Initializing the Model

The optimization criterion for our model is a complex non-convex function in terms of the transformations $T$ and part labels $\alpha$. Our hard EM algorithm is only capable of getting to a local minimum of this function. Therefore, it is dependent on a good starting point. This section addresses the problem of providing the EM algorithm with a good starting point.

### 6.1 Obtaining Transformation Estimates

One way of initializing the algorithm is by performing clustering in the space of rigid transformations, as suggested by Cheung *et al.* [8]. Since the correspondences between the template mesh $X$ and all instance meshes $Z_i$ are known, we can estimate the local rigid transformation between a point $x_j$ and its counterpart $z_{i,j}$. To do so, we look at small local patches centered at $x_j$ and $z_{i,j}$, and assume that the local transformation between the patches is rigid. Using ICP [6], the optimal rigid transformation $\tau_{i,j}$ registering these patches can be computed. Every $\tau_{i,j}$ can be represented as a vector in 6 dimensional Euclidean space. Each point $x_j$ becomes associated with $N$ such vectors, corresponding to its transformation in each instance mesh. The resulting stacked vectors can then be clustered by applying adaptive PCA [3], a variant of Gaussian Mixture Modeling. The cluster labels serve as an initial set of part labels to the points $x_j$. A result of this step is demonstrated in Fig. 1(B). As it does not exploit

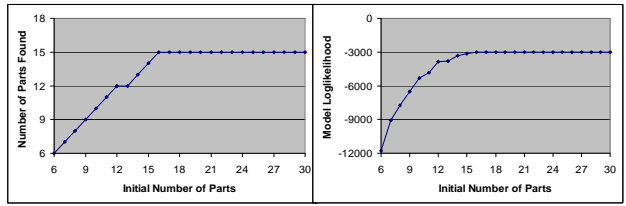

Figure 2: Graphs showing the log-likelihood and the number of parts of the final model using different number of parts as initialization, for the puppet dataset.

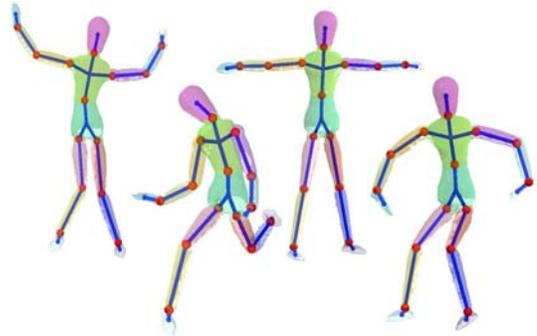

Figure 3: Four different poses from the puppet dataset display the 15 rigid parts and the articulated skeleton, both of which are recovered automatically

the connectivity of the mesh surface, it can serve as initialization to our main algorithm, but is not good enough by itself.

Using a Matlab implementation of adaptive PCA available on the web [3], clustering a set of 7 puppet poses (4000 points each) into 15 rigid parts takes about an hour on a Sun Blade 2000 dual-processor machine. Surprisingly, this pre-processing step becomes the bottleneck of the whole part-finding pipeline. Here we propose a more efficient way of initializing the model that gives comparable or even better results than clustering.

The main insight we will exploit is that the soft contiguity constraints introduce a preference for models which have fewer parts: The more parts there are, the more edges there are between mesh parts, the larger the penalty introduced by the pairwise contiguity potentials. Thus, we can start the model with a large number of possible parts, and redundant part hypotheses will be automatically pruned.

We therefore initialize the model by dividing the mesh into small surface patches, all of which have approximately the same area. This is done by uniformly subsampling the mesh, and assigning each point on the original mesh to the nearest point on the subsampled mesh. All points on the original mesh that are given the same assignment are grouped together to form a patch. This process takes a fraction of a second, compared to an hour for our previous initialization scheme.

When the subdivision into patches is fine enough, some rigid parts will contain patches that lie completely inside them, and the transformations for those patches from the model mesh $X$ to the morphed meshes $Z$ will closely approximate the corresponding transformations for the actual rigid parts. Using the patches as initial part assignments for our algorithm, we get a good starting point for the first M-Step.

### 6.2   Determining the Number of Parts

The idea of initializing the algorithm by subdividing the surface of mesh $X$ into patches leads to the question of how many initial patches are necessary. In principle, it is sensible to choose an initial number of patches $P$ to be larger than the actual number of rigid parts we expect. The larger $P$ is, the more likely it is to get a patch that lies completely inside a rigid part. As we discussed, our model encodes a preference for having fewer parts, so that redundant part hypotheses are pruned automatically. Indeed, our experiments (Fig. 2) show that, initially, as we increase the number of model parts $P$, the number of selected parts increases; but once the optimal number of parts is reached, increasing $P$ further does not increase the number of parts found.

In the case of rigid objects, the final number of parts found by our algorithm is generally the correct number of parts in the articulated objects. In cases of objects that also undergo non-rigid deformations, the number of parts found depends on the tradeoff between allowing more deformation within a part and splitting into more parts to preserve part rigidity. These preferences depend on the edge potential $\tau$ and the variance $\sigma^2$. As their ratio $\delta = \sigma^2/\log(\tau)$ increases, we allow more local deformation, where instance mesh points deviate from their predicted locations.

It turns out that, for large values of $\delta$, the problem becomes underconstrained, with multiple possible solutions that are plausible and have similar scores. This large hypothesis space makes the relaxed integer program considerably more difficult to solve, especially in the absence of good transformation estimates. To address the problem, we start with a low $\delta$ ratio, and gradually increase it as we iterate. The intuition behind this approach is that we separate the error due to random initialization from the error due to non-rigidity. During the early stages of the algorithm, there is a great deal of error due to random initialization; we therefore start with a smaller-than-desired $\delta$, heavily penalizing discrepancies from the rigid part assumption. As a side effect, our algorithm will tend to split a non-rigid part into several rigid parts, resulting in more parts than we want. As the algorithm con-verges, the noise from random initialization becomes less significant, so we can gradually relax the rigidity assumptions and anneal the value of $\delta$; this process results in the merging (and modification) of parts, and the elimination of unnecessary part hypotheses.

## 7 Learning an Articulated Object Skeleton

Once we obtain the part labels for every point in a mesh, it is easy to recover the joint between two adjacent rigid parts. We adapt the solution by Cheung et al. [8]. Suppose we want to find the joint between two adjacent parts $p$ and $q$. Let the coordinates of the joint in the model mesh be $y_{p,q}$. Since the joint belongs to two object parts simultaneously, it should satisfy the equation:

$$T_{ip}(y_{p,q}) = T_{iq}(y_{p,q}), \qquad i = 1 \ldots N \quad (9)$$

Putting together the equations for all instance meshes, $y_{p,q}$ is the solution to the following linear regression problem:

$$\operatorname*{argmin}_{y_{p,q}} \sum_{i=1}^{N} \|T_{ip}(y_{p,q}) - T_{iq}(y_{p,q})\|^2 \quad (10)$$

and can be solved using SVD.

Sometimes, the solution to the above equation can be an entire subspace of points. Suppose the joint only allows one degree of movement, such as the knee joint of a human leg. Then any point on the line perpendicular to the plane of allowed movement is a solution to the above linear regression problem.

We choose to resolve this problem by introducing an additional regularization term, which prefers that the joint position is close to the centroid of the points on the boundary between the two parts. After all, we want the joint to be close to where the two parts meet. More formally, denote the 'boundary centroid' in mesh $Z_i$ by $c_{p,q}^i$. Then $y_{p,q}$ is the solution to the following weighted linear regression problem:

$$\operatorname*{argmin}_{y_{p,q}} \sum_{i=1}^{N} \|T_{ip}(y_{p,q}) - T_{iq}(y_{p,q})\|^2$$
$$+ \gamma \sum_{i=1}^{N} \|\frac{1}{2}(T_{ip}(y_{p,q}) + T_{iq}(y_{p,q})) - c_{p,q}^i\|^2 \quad (11)$$

## 8 Experimental Results

We applied our algorithm to meshes from two different datasets. In one data set, we used a range scanner [11] to acquire a set of seven different complete surface meshes of a wooden puppet in different positions. Each mesh was constructed from ten range

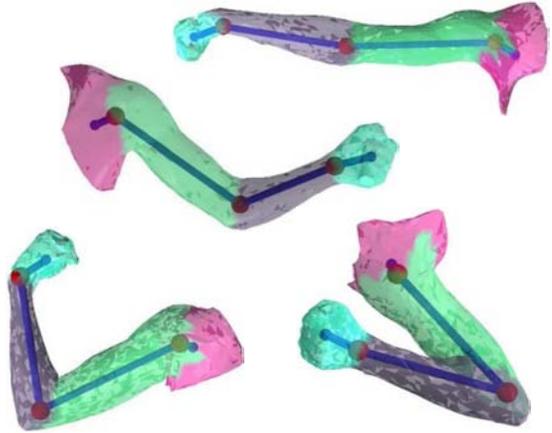

Figure 4: Four different poses from the arm dataset display four (approximately) rigid parts and the articulated skeleton, both of which are recovered automatically

scans taken from different viewing angles, composed using the method of Curless and Levoy [5], and subsampled to contain ~4000 points and 8000 triangles. Our second data set consisted of eight meshes of a human arm, acquired and used by Allen et al. [1]. These meshes are not complete surfaces, and so cannot be used directly by our algorithm; fortunately standard hole filling techniques (e.g., [10, 19]) can be used to construct a complete surface.

We automatically align one puppet mesh to the remaining six meshes in our puppet dataset using the Correlated Correspondences algorithm [2]. We experiment with both initialization approaches described in Sec. 6.1. Results shown in Fig. 1 demonstrate that both initialization methods performed equally well. However, the method where we initialize the M-step by partitioning the mesh $X$ into small surface patches is preferable because of its simplicity and overwhelming computational advantage. The correct model containing 15 parts was found whenever the number of surface patches in the initialization was equal to or greater than 16 (Fig. 2). More instances of the final model superposed onto the recovered articulated skeleton are displayed in Fig. 3. To our knowledge, this is the first implementation that estimates such a complex skeleton from real world data with very few poses, in a completely unsupervised way. For a visualization of the part-finding procedure (among other applications of the Correlated Correspondence algorithm), please refer to the movie at http://robotics.stanford.edu/~drago/cc/video.mp4.

The arm data is more challenging: the arm undergoes significant deformations as it bends, so that it is not purely an articulated model composed of rigid parts. Fig. 5 demonstrates the progress of our algorithm as the parameter $\delta$ is increased, until we end up with four parts, which is the intuitively correct number of parts for the arm (see Fig. 4). The partition of the object depends on the exact setting of the parameter $\delta$ (see Fig. 5 D and E). Setting the value of $\delta$ to be too large over-emphasizes the soft contiguity constraints. The part boundaries are shifted to a configuration minimizing the number of links between parts, ignoring the underlying geometric structure (Fig. 5 E). Our results on the arm dataset suggest that even in the presence of significant non-rigidity, such as twisting of the forearm and bulging of the biceps, our algorithm performs quite well.

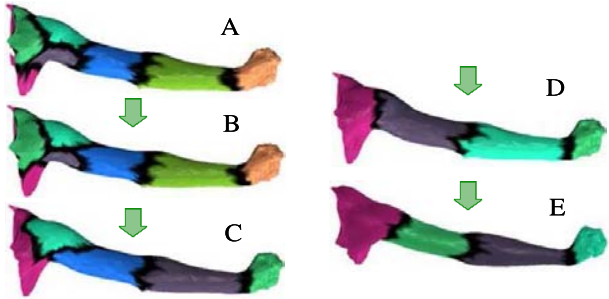

Figure 5: Illustration of annealing on the Arm dataset. The sequence above is obtained by starting with low $\delta$, and gradually increasing it after each iteration of EM, until our desired $\delta$ is reached. The algorithm anneals part hypotheses and eventually converges at four parts (D). Setting the value of $\delta$ to be too large (E) focuses more on the soft contiguity constraints and less on the underlying geometric structure. This results in partitions which reduce the number of links between parts.

## 9 Conclusion

We describe an algorithm which automatically recovers articulated object models given a set of 3D meshes of the object in different configurations. The algorithm first registers all the given meshes using the Correlated Correspondence [2] algorithm. Then it iteratively estimates the part assignment for each point and the rigid transformation of each part. Once the part assignments are recovered, the joints are estimated by articulation constraints.

We apply the algorithm to two challenging real-world datasets, one having a large number of rigid parts, and one consisting of parts that are slightly deformable. In both cases the algorithm recovers their articulated models in an unsupervised manner from only a small number of meshes. Our algorithm not only recovers the parts and joints, but also figures out the optimal number of parts automatically.

In this paper, we have decoupled the registration algorithm from the algorithm which recovers the articulated object structure. While ideally both steps could be executed simultaneously, this decoupling allows us to apply robust global inference strategies during the registration process and during the inference step partitioning the object surface into parts. The ability to perform robust and efficient global inference is very important, because it helps us to circumvent many local maxima during both processes. Our approach can be bootstrapped in a fairly straightforward way to use the computed rigid parts and their transformations to improve the registration results. However, there was little to be gained from such bootstrapping on these data sets given the quality of our registration results.

There are many interesting directions in which this work can be extended. Most obviously, we would like to introduce into our object model the ability for parts to have small shape deformations, while still preserving the assumption that large deformations occur only in articulated joints. It would also be interesting to automatically learn a model of the allowable deformations at different joints. More long-term, it would be interesting to explore the use of these ideas to important tasks such as a marker-free solution to the limb-tracking problem in medical and other applications.

### Acknowledgements

We thank Ben Taskar for useful discussions. This work has been supported by the Office of Naval Research, under Young Investigator (PECASE) grant N00014-99-1-0464, and ONR Grant N00014-00-1-0637 under the DoD MURI program.